\documentclass[10pt,twocolumn,letterpaper]{article}

\usepackage{cvpr}              

\usepackage{cuted}
\usepackage{algorithm,algpseudocode}

\usepackage{float}

\definecolor{cvprblue}{rgb}{0.21,0.49,0.74}
\usepackage[pagebackref,breaklinks,colorlinks,allcolors=cvprblue]{hyperref}

\def\paperID{*****} 
\def\confName{CVPR}
\def\confYear{2026}

\title{SDFoam: Signed-Distance Foam for explicit surface reconstruction}

\author{
\textbf{Antonella Rech}$^{1}$\quad
\textbf{Nicola Conci}$^{1,2}$\quad
\textbf{Nicola Garau}$^{1,2}$\\[0.5em]
$^1$University of Trento, Italy \quad
$^2$CNIT  \\
\url{https://mmlab-cv.github.io/SDFoam}
}

\begin{document}
\maketitle


\begin{strip}
    \centering
    \includegraphics[width=.7\textwidth]{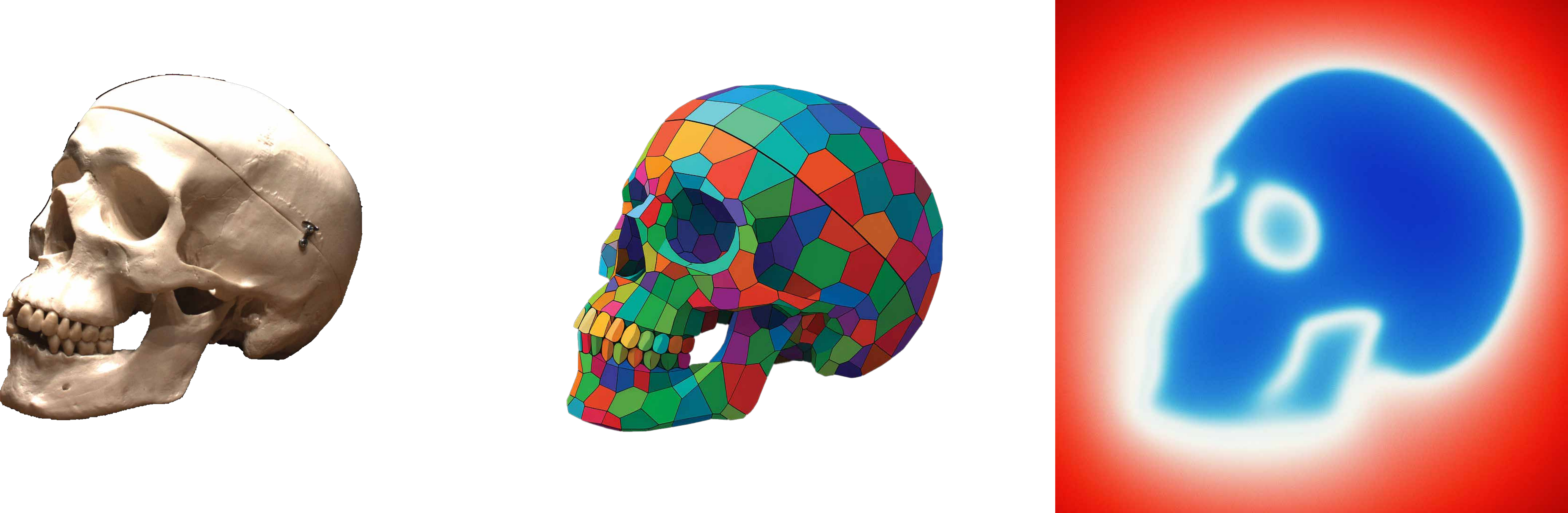}
    \captionof{figure}{Existing methods in literature reconstruct 3D scenes either by employing explicit or implicit geometry, each with their own advantages and drawbacks. Our method, \textit{SDFoam}, jointly learns a signed distance field (\textit{SDF}) and a 3D Voronoi Diagram (or \textit{foam}), both of which are optimized during a ray-tracing process. Our method offers a good trade-off of rendering speed, visual fidelity and reconstruction accuracy. Our code and experiments are available at \url{https://mmlab-cv.github.io/SDFoam}.}
    \label{fig:teaser}
\end{strip}

\begin{abstract}
Neural radiance fields (NeRF) have driven impressive progress in view synthesis by using ray-traced volumetric rendering. Splatting-based methods such as 3D Gaussian Splatting (3DGS) provide faster rendering by rasterizing 3D primitives. RadiantFoam (RF) brought ray tracing back, achieving throughput comparable to Gaussian Splatting by organizing radiance with an explicit Voronoi Diagram (VD). Yet, all the mentioned methods still struggle with precise mesh reconstruction. We address this gap by jointly learning an explicit VD with an implicit Signed Distance Field (SDF). The scene is optimized via ray tracing and regularized by an Eikonal objective. The SDF introduces metric-consistent isosurfaces, which, in turn, bias near-surface Voronoi cell faces to align with the zero level set. The resulting model produces crisper, view-consistent surfaces with fewer floaters and improved topology, while preserving photometric quality and maintaining training speed on par with RadiantFoam. Across diverse scenes, our hybrid implicit-explicit formulation, which we name SDFoam, substantially improves mesh reconstruction accuracy (Chamfer distance) with comparable appearance (PSNR, SSIM), without sacrificing efficiency.
\end{abstract}    
\section{Introduction} 
\label{sec:intro}

Reconstructing 3D geometry and appearance from multi-view images is a long-standing problem in computer vision and graphics. Classical approaches based on multi-view stereo or surface meshing struggle to recover accurate geometry for scenes with complex materials, thin structures, or indirect illumination. Neural implicit representations have recently emerged as a powerful alternative, achieving remarkable reconstruction quality by learning the scene composition directly from images.

In particular, radiance-based methods \cite{mildenhall2020nerf, barron2021mipnerf, mueller2022instant, li2023neuralangelo} represent a scene as a volumetric density field optimized for appearance, whereas SDF-based methods \cite{wang2021neus, yariv2021volume, yariv2020multiview} employ signed distance functions to recover implicit surfaces with higher geometric accuracy. However, learning both visually and geometrically consistent scenes still remains challenging.

We propose \textbf{SDFoam}, a framework that unifies geometry reconstruction and appearance modeling within a single representation (Fig. \ref{fig:teaser}). As in Radiant Foam \cite{govindarajan2025radfoam}, we represent the scene as a 3D Voronoi diagram whose dual Delaunay triangulation induces a sparse volumetric mesh. In SDFoam, however, each Voronoi cell is parameterized not only by its centroid and color, but also by a locally defined signed distance value, turning the Voronoi–Delaunay structure into a jointly implicit–explicit representation that is both differentiable and spatially coherent. The signed distance field is learned together with the Voronoi structure through differentiable rendering, allowing the same representation to drive both surface reconstruction and view synthesis.

A key contribution of SDFoam is its novel, fast mesh extraction strategy. Instead of relying on post-hoc surface reconstruction algorithms, our method extracts the surface directly from the trained Voronoi structure by leveraging the SDF. This allows a direct transition between implicit and explicit representations without altering the learned surface topology. 


In summary, SDFoam:
\begin{itemize}
    \item Offers a good trade-off between visual appearance and mesh reconstruction, closing the gap between radiance fields and SDF-based methods;
    \item Accelerates SDF-based reconstruction and rendering by exploiting a dual Voronoi-Delaunay structure, yielding faster training and inference than prior SDF approaches;
    \item Is, to our knowledge, the first framework to couple a 3D Voronoi tessellation with a learned SDF, providing a jointly implicit-explicit scene representation;
    \item Enables fast mesh extraction, up to 5$\times$ faster than a naive density-based thresholding on RadiantFoam \cite{govindarajan2025radfoam}.
\end{itemize}



\section{Related Work} 
\label{sec:related}

\subsection{View synthesis}

View synthesis aims to generate novel views of a scene given a set of observed images. Early approaches relied on classical reconstruction methods such as Structure-from-Motion (SfM)~\cite{snavely2006photo}, to obtain coarse reconstructions via sparse point clouds, and Multi-View Stereo (MVS), which can guide image re-projection for novel view synthesis, through a denser 3D reconstrution \cite{goesele2007multi}. 
These methods suffer from inherent limitations: regions with missing or poorly observed data result in incomplete reconstructions, and over-reconstruction can occur in areas with uncertain geometry. Neural rendering approaches address these limitations by learning continuous and differentiable scene representations that enable inference of missing regions, recovery of fine geometric details, and synthesis of novel views directly from images.

Neural Radiance Fields (NeRF)\cite{mildenhall2021nerf} marked a significant breakthrough in neural rendering and novel view synthesis. NeRF models a scene as a continuous volumetric field of density and radiance, parameterized by a neural network. Although NeRF achieves high-quality results, it is computationally intensive, and many subsequent methods have focused on accelerating both training and rendering. For instance, Instant-NGP\cite{mueller2022instant} leverages a hash table-based data structure to improve efficiency, while Plenoxels\cite{yu2022plenoxels} represents the scene as a sparse voxel grid of spherical harmonics, eliminating the need for a neural network. Other notable approaches, such as TensoRF\cite{Chen2022ECCV}, DVGO\cite{SunSC22}, FastNeRF\cite{mildenhall2020nerf}, Mip-NeRF\cite{barron2021mipnerf}, and SparseNerF\cite{wang2023sparsenerf}, also employ different strategies to accelerate computation while maintaining visual quality.

Despite achieving high visual quality, in large part they still suffer from dense sampling along rays. 
This motivated the development of point-based methods, which represent a scene as sparse collections of points or primitives, where each primitive encodes local geometry and appearance, reducing unnecessary sampling.

Building on the idea of point-based methods, Gaussian Splatting (3DGS)\cite{kerbl3Dgaussians} models each scene element as an anisotropic 3D Gaussian with color and view-dependent properties. Beyond 3DGS, several alternative 3D and 2D primitive representations have been explored. 2DGS\cite{Huang2DGS2024} uses 2D Gaussians, Triangle Splatting\cite{Held2025Triangle} represents scenes with explicit surface triangles, and methods such as DMTet\cite{shen2021deep}, Deformable Beta Splatting\cite{liu2025deformablebetasplatting}, and Tetrahedron Splatting\cite{gu2024tetrahedron} leverage tetrahedral or deformable Beta primitives.

RadiantFoam\cite{govindarajan2025radfoam} introduced a novel approach that represents the scene using a Voronoi Diagram and Delaunay triangulation. Extending the concept of point-based methods, RadiantFoam models the scene with non-overlapping polyhedral primitives, each capturing local geometry and appearance. By combining the flexibility of point-based representations with such Voronoi-based scene, the approach enables accelerated volume rendering through ray tracing.

Despite significant progress in view synthesis, most approaches in the literature focus on rendering quality rather than accurate geometry reconstruction, limiting their applicability to tasks that require precise surface modeling.

\subsection{Mesh Reconstruction}

Rather than focusing on accurate surface recovery, many relegate geometric reconstruction to a secondary role, privileging view synthesis. Mesh reconstruction aims at recovering surface representations of the scene, prioritizing geometric accuracy. Several methods have been proposed for this task, such as SuGAR~\cite{guedon2023sugar}, VoroMesh~\cite{maruani23iccv}, and TSDF Fusion~\cite{zeng20163dmatch}.

More recent works aim to obtain both view synthesis and geometry from a single representation, notably IDR~\cite{yariv2020multiview}, UNISURF~\cite{Oechsle2021ICCV}, Neuralangelo~\cite{li2023neuralangelo}, NeuS~\cite{wang2021neus}, VolSDF~\cite{yariv2021volume}, and NeRF2Mesh~\cite{tang2022nerf2mesh}. These methods typically leverage implicit surface representations, such as SDFs, combined with differentiable rendering, enabling the joint optimization of photorealistic appearance and accurate 3D geometry. However, they still often rely on post-hoc mesh extraction algorithms, such as Marching Cubes~\cite{lorensen1998marching}, Poisson Reconstruction~\cite{kazhdan2006poisson}, Dual Contouring~\cite{ju2002dual}, Marching Tetrahedra~\cite{doi1991efficient}, or Ball Pivoting~\cite{bernardini2002ball}, which often struggle with numerical approximations and can compromise geometric precision.

In contrast, we propose a novel framework that leverages the Voronoi diagram’s spatial structure, similar to \cite{govindarajan2025radfoam}, and differentiable ray tracing to jointly learn geometry and appearance. Each Voronoi cell is modeled as a local SDF, allowing primitives to adaptively align with the underlying scene geometry. Unlike previous works, our method directly extracts faces from the Voronoi topology, thus providing a much faster (up to 5$\times$) mesh reconstruction speed while preserving the reconstruction accuracy.

\begin{figure}
  \centering
  \includegraphics[width=.4\textwidth]{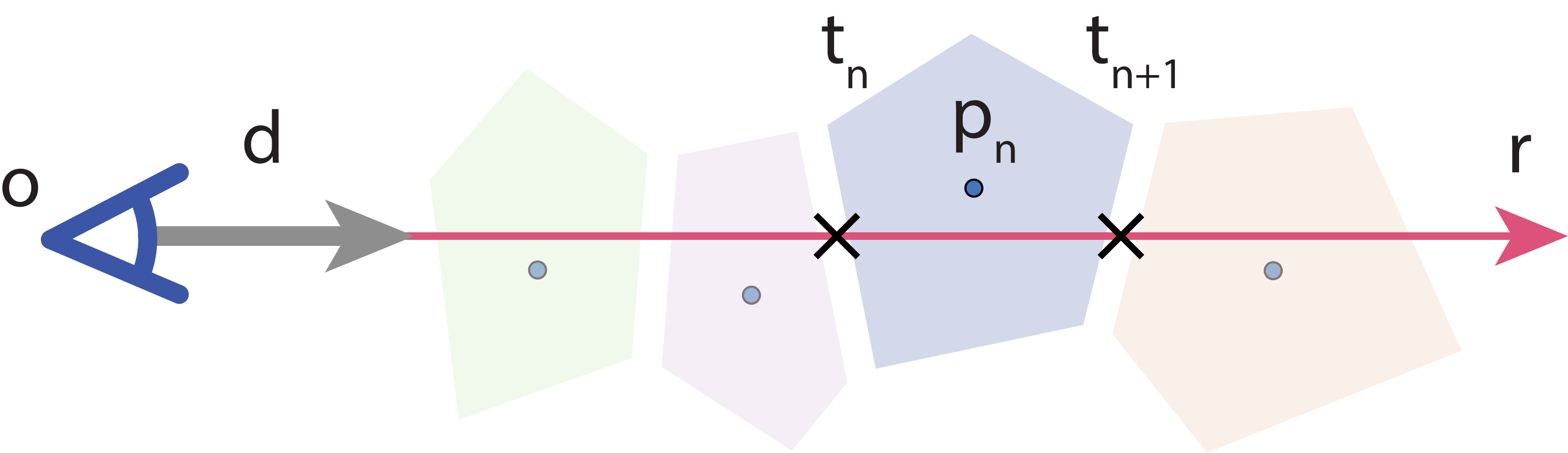}
  \captionof{figure}{Ray traversal through Voronoi cells. The ray intersects the $n$-th cell (centered at site $p_n$) at positions $t_n$ (entry) and $t_{n+1}$ (exit), defining the segment length $\delta_n$. Spatial and visual information are piecewise constant within $\delta_n$. The ray $r$ is defined by its origin $o$ and direction $d$.}

  \label{fig:sdf_voronoi}
\end{figure}

\begin{figure*}
  \centering
  \includegraphics[width=\textwidth]{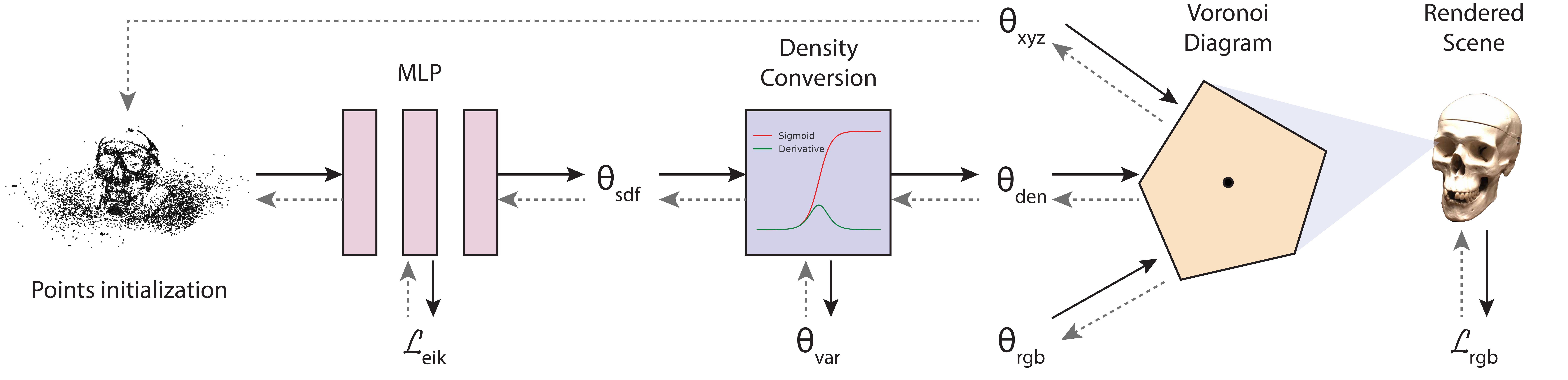}
  \captionof{figure}{SDFoam Architecture. A point cloud is initialized and refined over time by learning an SDF from its points. Their SDF values are then converted to density, which jointly with color and position parameters are used to learn a ray-traced scene. $\theta_{var}$ is a learnable variance parameters that allows to improve the SDF-to-density conversion over time, similar to \cite{wang2021neus}.}
  \label{fig:architecture}
\end{figure*}

\section{Method} 
\label{sec:method} 

Our goal is to develop a unified representation that simultaneously captures both geometry and appearance. In SDFoam, we jointly learn an SDF and a 3D Voronoi diagram, where the Voronoi cells serve as primitives for both tasks (Figure~\ref{fig:architecture}). Each cell is defined by local parameters that encode its spatial properties, such as centroid position and SDF value, and visual properties, such as color and spherical harmonics. During training, the centroids are optimized to reconstruct the SDF from multi-view images and to model the color and light of the scene through differentiable rendering. This dual representation enables the same Voronoi structure to act as a unified framework for geometry and view synthesis, while naturally yielding a mesh-expressible surface, thereby avoiding the artifacts, numerical errors, and topological ambiguities typically associated with discretization-based implicit surface extraction methods (e.g., Marching Cubes\cite{lorensen1998marching}, Poisson Reconstruction\cite{kazhdan2006poisson}, etc.).

In the following sections, we outline our approach, covering the formulation of the Voronoi Diagram (Section~\ref{sec:voronoi}), the transformation from SDF to volume density (Section ~\ref{sec:sdf2density}), the differentiable rendering procedure based on ray tracing (Section ~\ref{sec:diff_ray_tracing}), the joint optimization of geometry and appearance (Section ~\ref{sec:optimization}), the Prune and Densify strategy (Section ~\ref{sec:densify}), and the mesh extraction procedure (Section ~\ref{sec:mesh}).

\subsection{Voronoi‑based SDF representation}
\label{sec:voronoi}
Let $\mathcal{P}=\{\mathbf{p}_i\}_{i=1}^{N}$ be a set of optimizable
\emph{sites} in $\mathbb{R}^3$. The Voronoi diagram partitions the space into a set of convex cells ${\mathcal{C}_i}$, each associated with a site (or generator) $\mathbf{p}_i \in \mathbb{R}^3$. Intuitively, each cell contains all points that are closer to its corresponding site than to any other. Formally, the $i^{\text{th}}$ Voronoi cell is defined as:

\begin{equation}
\mathcal{C}_i = \bigl\{ \mathbf{x} \in \mathbb{R}^3 \;\big|\; 
\|\mathbf{x} - \mathbf{p}_i\| \le \|\mathbf{x} - \mathbf{p}_j\|, \ \forall j \neq i \bigr\},
\end{equation}
where $\mathbf{x}$ denotes a point in the 3D space and $\mathbf{p}_j \in \mathcal{P}$ represents any site other than $\mathbf{p}_i$.

The advantage of using the Voronoi diagram lies in its dual formulation — the Delaunay triangulation.
In this dual structure, each pair of sites whose Voronoi cells share a common face are connected by an edge, forming a tetrahedral mesh that covers the convex hull of $\mathcal{P}$.
Formally, the Delaunay triangulation $\mathcal{D}(\mathcal{P})$ is defined as the set of tetrahedra whose circumspheres contain no other sites in their interior:

\begin{equation}
\mathcal{D}(\mathcal{P}) = 
\bigl\{\, T \;\big|\; \Omega(T) \cap (\mathcal{P} \setminus T) = \emptyset \,\bigr\}.
\end{equation}
\\
where $T = (\mathbf{p}_1, \mathbf{p}_2, \mathbf{p}_3, \mathbf{p}_4) \subset \mathcal{P}$ is the tethahedron and
and $\Omega(T)$ denotes the circumsphere of $T$.

Building on this dual structure, we represent the scene using a Voronoi diagram, similar to \cite{govindarajan2025radfoam}, but differing in the type of information encoded within each cell. 

In our formulation ($\mathcal{P} = \{\mathbf{p}_i\}_{i=1}^{N}$), each site $\mathbf{p}_i$ is associated with a set of parameters:
\begin{itemize}
    \item a density value $\rho_i$, obtained by evaluating the global signed distance function $f_\theta(\mathbf{x})$ at the cell centroid $\mathbf{p}_i$ and transforming it via the mapping $\phi_s$ described in Section~\ref{sec:sdf2density}
    \item a color vector $\mathbf{c}_i \in \mathbb{R}^3$,
    \item spherical harmonic coefficients $\mathbf{h}_i$ encoding view-dependent appearance
    \item the centroid position $\mathbf{p}_i$ of the corresponding Voronoi cell $\mathcal{C}_i$.
\end{itemize}

Each parameter is treated as piecewise constant within its Voronoi region $\mathcal{C}_i$, enabling a fast and efficient differentiable ray-tracing algorithm as described in Section~\ref{sec:diff_ray_tracing}.

\subsection{From SDF to Density}
\label{sec:sdf2density}

To reconstruct a 3D surface from a set of 2D images, we represent the scene using a global neural implicit signed distance function (SDF) $f_\theta(\mathbf{x})$, whose zero-level set defines the surface:
\begin{equation}
\mathcal{S} = \{\mathbf{x} \in \mathbb{R}^3 \mid f_\theta(\mathbf{x}) = 0 \}
\end{equation}

For each Voronoi cell $\mathcal{C}i$, we associate a density value $\rho_i$, obtained by evaluating the SDF $f\theta(\mathbf{x})$ at the cell centroid $\mathbf{p}i$ and transforming it through the mapping function $\phi_s$:
\begin{equation}
\rho_i = \phi_s(f\theta(\mathbf{p}_i)),
\end{equation}
which assigns a probability-like value that represents the contribution of the cell to the final pixel color.

Using the per-cell parameters, differentiable volume rendering accumulates colors and opacities along camera rays to generate synthetic images.
These images are then compared with the ground-truth views, and the resulting gradients are backpropagated to update the SDF network parameters (Figure \ref{fig:architecture}).
This process effectively bridges the implicit 3D representation and the 2D supervision, enabling the joint learning of accurate surface geometry and view-dependent appearance.

Directly interpreting the SDF as a density function introduces bias in the reconstructed surfaces, as discussed in NeuS~\cite{wang2021neus}.  
Following their formulation, we define the mapping function \(\phi_s(f(\mathbf{x}))\) as the derivative of the sigmoid:

\begin{equation}
\phi_s(f) 
\;=\; \frac{d}{df} \sigma(f)
\;=\; \beta \, \sigma(f) \,(1 - \sigma(f)),
\label{eq:sdf_density_math}
\end{equation}

where the sigmoid function is defined as

\begin{equation}
\sigma(f) = \frac{1}{1 + e^{-\beta f}},
\label{eq:sigma}
\end{equation}

with \(f\) being the SDF value and \(\beta\) the trainable sharpness parameter.

Intuitively, $\phi_s(f)$ attains its highest values near the zero-level set of the SDF and decays symmetrically away from the surface. The sharpness parameter $\beta$ controls the spread of this bell-shaped distribution: larger values of $\beta$ produce narrower peaks (lower variance), resulting in sharper and more refined surface.

This mapping satisfies two key properties essential for surface reconstruction: (1) it assigns maximal weight to the zero-level set, ensuring that color contributions predominantly originate from the surface, and (2) it remains fully differentiable with respect to $f_\theta$ and $\beta$, enabling gradient-based optimization of both geometry and density mapping.

\subsection{Differentiable ray tracing and rendering}
\label{sec:diff_ray_tracing}

Rendering a pixel requires integrating radiance along its camera ray
\begin{equation}
\mathbf{r}(t) = \mathbf{o} + t \mathbf{d},
\end{equation}
where $\mathbf{o}$ is the camera center and $\mathbf{d}$ is a unit direction.

Following RadiantFoam\cite{govindarajan2025radfoam}, we traverse the sequence of Voronoi cells intersected by a ray and accumulate colors and opacities piecewise. Each ray is subdivided into segments corresponding to the cells it intersects. Let $t_n$ and $t_{n+1}$ denote the entry and exit positions of the ray on the cell for the $n$-th segment, and define the segment length as
\begin{equation}
\delta_n = t_{n+1} - t_n.
\end{equation}
Figure~\ref{fig:sdf_voronoi} illustrates how the ray interacts with the Voronoi cell.

Within each segment, the density $\rho_n$ and radiance $\mathbf{L}_n(\mathbf{d})$ are assumed constant. The opacity of the segment is
\begin{equation}
\alpha_n = 1 - \exp(-\rho_n \, \delta_n),
\label{eq:alpha}
\end{equation}
where $\rho_n$ is the density obtained from the SDF evaluated at the centroid
and the accumulated transmittance along the ray up to the segment is
\begin{equation}
T_n = \prod_{k=1}^{n-1} (1 - \alpha_k).
\end{equation}

The contribution of each segment to the pixel radiance is then
\begin{equation}
\mathbf{L} = \sum_n T_n \, \alpha_n \, \mathbf{L}_n(\mathbf{d}).
\label{eq:rendering}
\end{equation}

This piecewise constant formulation can be seen as an approximation of the continuous volume rendering integral, since the density \(\rho_n\) and radiance \(\mathbf{L}_n(\mathbf{d})\) are assumed constant within each Voronoi cell segment and the segment widths \(\delta_n\) are computed from exact ray–cell intersections. By leveraging these intersections, \(\delta_n\) varies continuously with the cell positions, enabling fully differentiable ray tracing. Gradients can thus propagate to update the SDF network, densities \(\rho_n\), radiance \(\mathbf{L}_n\), and spherical harmonic coefficients.

 It is important to note that the same Voronoi cell can be traversed by multiple rays. To maintain consistent density values across views, we model the density as a parameter of the Voronoi cell itself, rather than as a function along a ray representing the probability of hitting the surface, as in NeuS. We then apply the mapping formula~\eqref{eq:sdf_density_math} at the centroid of each cell and calculate alpha using equation ~\eqref{eq:alpha}.

\subsection{Optimization and regularization}
\label{sec:optimization}
The parameters to be learned include the site positions $\mathbf{p}_i$,
SDF values $f_i$, the color $\mathbf{c}_r$, the sh coefficients $h_i$ and the sharpness $\beta$ of the sigmoid function. Optimization proceeds by minimizing the following loss
\begin{equation}
 \mathcal{L}oss 
 = \mathcal{L}_{\mathrm{rgb}} + \lambda_{\mathrm{eik}} \, \mathcal{L}_{\mathrm{eik}},
 \label{eq:loss}
\end{equation}

where the photometric loss \(\mathcal{L}_{\mathrm{rgb}}\) compares rendered pixel colors to ground-truth images, and we adopt an \(L2\) formulation as in Radiant Foam. This loss also supervises the SDF MLP, since the density of each Voronoi cell is computed from the corresponding SDF value. 

Moreover, we introduce an \emph{Eikonal} regularizer on the SDF gradient. The Eikonal loss encourages the norm of the SDF gradient to be close to 1 everywhere, which ensures that the distance changes linearly with position in space. Without this regularization, the SDF predicted by the MLP may present gradients that are either too small or too large, leading to distorted surfaces and biased density values for volume rendering.

In our case, the gradient is defined at the centroids:
\begin{equation}
\mathcal{L}_{\mathrm{eik}}
= \frac{1}{N} \sum_{i=1}^{N} \left( \|\nabla f_\theta(\mathbf{p}_i)\| - 1 \right)^2,
\end{equation}

where \(N\) is the number of Voronoi cells, \(\mathbf{p}_i\) is the centroid of the \(C_i\) cell, and \(\nabla f_\theta(\mathbf{p}_i)\) is the gradient of the SDF evaluated at that centroid. The weight is set to \(\lambda_{\mathrm{eik}} = 0.01\).

\subsection{Pruning and Densification}
\label{sec:densify}
During optimization, the Voronoi diagram representation can be modified by adding or removing sites. The idea is similar to the splitting procedure used in \cite{kerbl3Dgaussians}, which aims to achieve high-detail view synthesis.

Voronoi cells with large photometric contributions are selected for splitting into smaller cells by applying small perturbations to the centroid positions. The new cells inherit the SDF and radiance parameters from the parent cell. Conversely, small cells with low rendering contribution are pruned. These operations are performed periodically during training and require rebuilding the Voronoi diagram each time they are applied.

\paragraph{Training protocol}

We implement SDFoam in PyTorch with custom CUDA kernels for ray tracing and the Voronoi representation. The Adam optimizer is used to minimize Eq.~\eqref{eq:loss}. The optimization uses separate learning rates for different parameter groups. 

Voronoi site positions are updated with an initial learning rate of $2\times10^{-4}$, decayed to $5\times10^{-6}$ using a cosine annealing schedule. SDF parameters are updated with an initial learning rate of $5\times10^{-4}$, decayed to $5\times10^{-5}$, and the sharpness parameter $\beta$ is updated with a fixed learning rate of $0.05$. Color and spherical harmonic coefficients start with a learning rate of $5\times10^{-3}$ and decay to $5\times10^{-4}$.

During an initial warm-up phase, the number of Voronoi sites is progressively increased until a predefined maximum is reached.
\subsection{Mesh Extraction}
\label{sec:mesh}

To extract a mesh from the trained Voronoi-based scene, we first filter the Voronoi cells using the learned SDF. Specifically, we retain only the cells whose centroids lie sufficiently close to the zero-level set of the SDF, effectively removing all cells far from the surface. This yields a set of volumetric Voronoi cells that surround the object’s surface. We observed that in our case the threshold give best results if set to 0.1.

\begin{figure*}
  \centering
  \includegraphics[width=\textwidth]{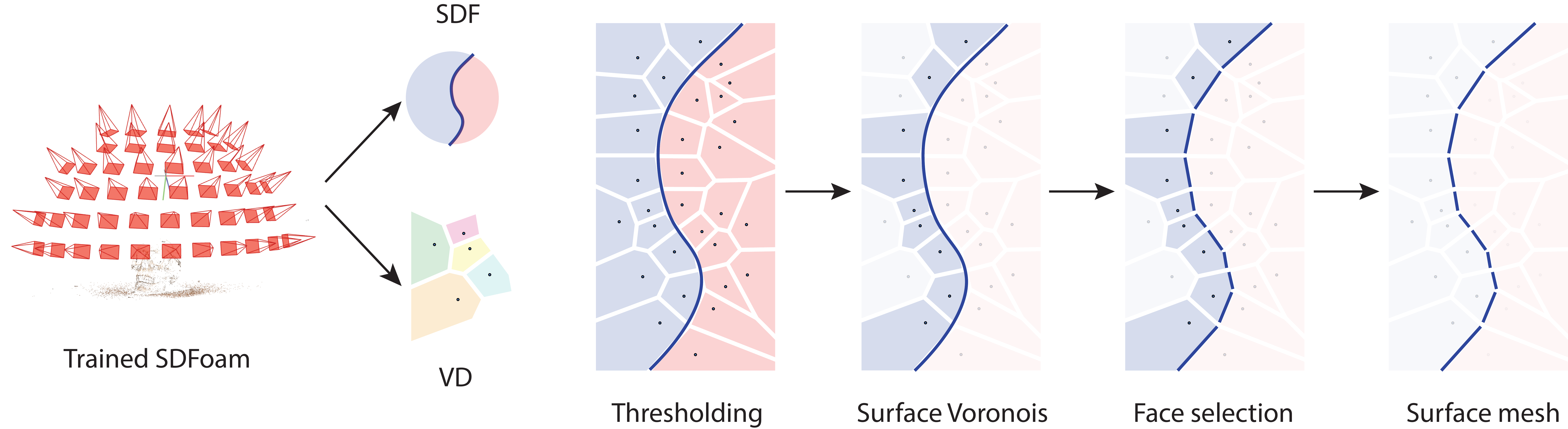}
  \captionof{figure}{From a trained SDFoam scene, we have access to both the SDF and the Voronoi Diagram. We infer the SDF value for each cell site, extracting the surface voronois via a threshold. The relevant surface faces are selected by thresholding their vertices against a close to zero SDF value. Since the VD is non-overlapping by nature, we don't need to build additional connectivity at this step.}
  \label{fig:mesh_reconstruction}
\end{figure*}

In the second step, we extract the faces of these filtered Voronoi cells to reconstruct the surface mesh. The procedure is straightforward: for each retained Voronoi cell, we keep only those faces that have at least three vertices with SDF values below a very small threshold, which we set to 0.001 in all our experiments. This ensures that only faces lying on or near the surface are included, while faces corresponding to interior or exterior regions are discarded. Figure~\ref{fig:mesh_reconstruction} illustrates this mesh extraction process.

\begin{figure*}
    \centering
    \includegraphics[width=\textwidth]{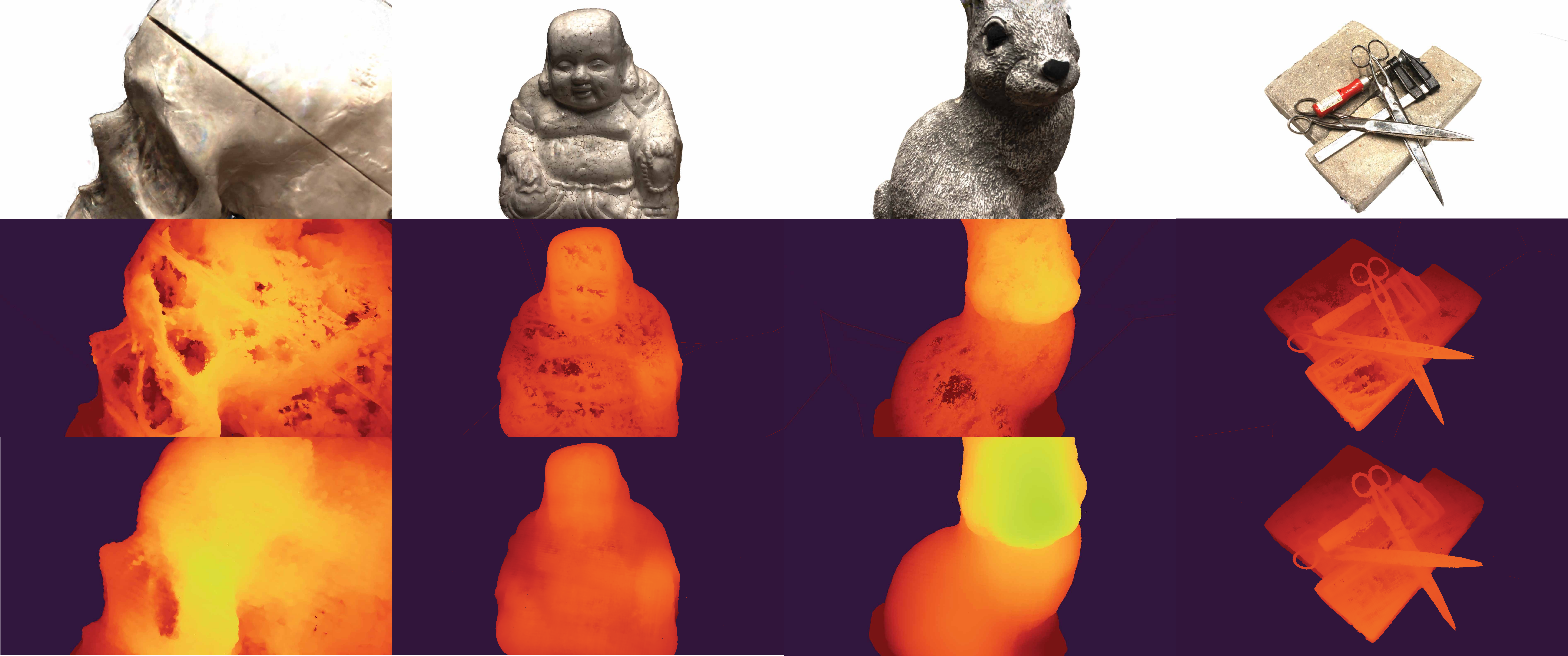}
    \captionof{figure}{Mesh reconstruction qualitative results. Top to bottom: ground truth, RF, SDFoam. Modelling the voronoi cells as local SDFs improves the consistency of the extracted surface, thus filling the typical \textit{holes} derived from the ray-tracing procedure in RF.}
    \label{fig:mesh_qualitative}
\end{figure*}

The main advantages of this approach are its simplicity and consistency. First, it does not require a separate mesh extraction algorithm, which typically introduces numerical approximations and topological artifacts. Instead, the mesh is obtained directly by selecting the relevant parts of the 3D Voronoi structure without altering its topology. Second, the extracted mesh is inherently correlated with the view synthesis, as each face already has associated color and texture information from the learned Voronoi cells. Finally, we can observe that the approach yields a high-quality mesh reconstruction, as can be seen in Fig. \ref{fig:mesh_qualitative}.

\begin{table*}[h]
\centering
\begin{tabular}{l|ccccc|cc|cccc}
\hline
 & \multicolumn{5}{c}{Chamfer Distance (w/ mask) $\downarrow$} & \multicolumn{2}{c}{PSNR $\uparrow$} & \multicolumn{2}{c}{SSIM $\uparrow$} \\
\hline
ScanID & IDR & NeRF & NeuS & RF* & SDFoam & RF & SDFoam & RF & SDFoam \\
\hline
scan24  & 1.63 & 1.83 & 0.83 & 6.13 & 1.86 & 31.28 & 29.80 & 0.877 & 0.848  \\
scan37  & 1.87 & 2.39 & 0.98 & 3.53 & 2.87 & 31.58 & 30.42 & 0.921 & 0.899  \\
scan40  & 0.63 & 1.79 & 0.56 & 6.02 & 1.80 & 32.12 & 30.68 & 0.886 & 0.851 \\
scan55  & 0.48 & 0.66 & 0.37 & 1.31 & 0.86 & 33.04 & 32.01 & 0.957 & 0.946 \\
scan63  & 1.04 & 1.79 & 1.13 & 7.10 & 2.23 & 35.97 & 35.26 & 0.969 & 0.962  \\
scan65  & 0.79 & 1.44 & 0.59 & 2.10 & 1.52 & 32.96 & 32.43 & 0.958 & 0.952 \\
scan69  & 0.77 & 1.50 & 0.60 & 2.84 & 1.30 & 29.18 & 28.17 & 0.937 & 0.907 \\
scan83  & 1.33 & 1.20 & 1.45 & 7.54 & 1.27 & 33.10 & 32.82 & 0.974 & 0.970  \\
scan97  & 1.16 & 1.96 & 0.95 & 5.41 & 1.53 & 31.25 & 30.12 & 0.974 & 0.931 \\
scan105 & 0.76 & 1.27 & 0.78 & 8.16 & 2.16 & 32.37 & 32.15 & 0.953 & 0.945 \\
scan106 & 0.67 & 1.44 & 0.52 & 6.79 & 1.67 & 29.08 & 28.41 & 0.948 & 0.931 \\
scan110 & 0.90 & 2.61 & 1.43 & 2.54 & 3.03 & 29.86 & 29.37 & 0.954 & 0.941 \\
scan114 & 0.42 & 1.04 & 0.36 & 1.78 & 1.15 & 32.28 & 31.01 & 0.946 & 0.930 \\
scan118 & 0.51 & 1.13 & 0.45 & 1.96 & 1.52 & 31.73 & 31.01 & 0.968 & 0.951 \\
scan122 & 0.53 & 0.99 & 0.45 & 1.82 & 1.34 & 34.82 & 34.12 & 0.976 & 0.969 \\
\hline
mean    & 0.90 & 1.54 & 0.77 & 4.33 & 1.74 & 32.04 & 31.18 & 0.947 & 0.929 \\
\hline
\end{tabular}
\caption{Quantitative evaluation on the DTU dataset. \textbf{*RF} $\rightarrow$ large Chamfer distances are due to floaters being difficult to filter with a naive density-based thresholding. Our method offers a good trade-off between mesh reconstruction and visual fidelity.}
\end{table*}

\begin{table*}[h]
\centering
\scriptsize
\begin{tabular}{l|cccccccccccccccccc|c}
\hline
\textbf{Scan ID} & 24 & 37 & 40 & 55 & 63 & 65 & 69 & 83 & 97 & 105 & 106 & 110 & 114 & 118 & 122 & Mean \\
\hline
PSNR ($NeRF$) & \textbf{24.83} & \textbf{25.35} & \textbf{26.87} & \textbf{27.64} & \textbf{30.24} & \underline{29.65} & \textbf{28.03} & \textbf{28.94} & \textbf{26.76} & \textbf{29.61} & \textbf{32.85} & \textbf{31.00} & \textbf{29.94} & \textbf{34.28} & \underline{33.69} & \textbf{29.31} \\
PSNR ($NeuS$) & 23.98 & 22.79 & \underline{25.21} & \underline{26.03} & \underline{28.32} & \textbf{29.80} & \underline{27.45} & \underline{28.89} & \underline{26.03} & \underline{28.93} & \underline{32.47} & \underline{30.78} & \underline{29.37} & \underline{34.23} & \textbf{33.95} & \underline{28.55} \\
PSNR ($RF$) & 22.60 & 21.42 & 22.83 & 24.50 & 24.90 & 22.49 & 22.96 & 25.03 & 21.33 & 25.74 & 29.05 & 28.84 & 25.03 & 30.34 & 25.55 & 24.84 \\
PSNR ($SDFoam$) & \underline{24.12} & \underline{22.87} & 21.67 & 22.36 & 26.71 & 23.38 & 24.59 & 24.32 & 22.87 & 20.80 & 26.56 & 27.70 & 25.95 & 26.20 & 24.27 & 24,30 \\
\hline
SSIM ($NeRF$) & 0.753 & \textbf{0.794} & \underline{0.780} & \underline{0.761} & \textbf{0.915} & \underline{0.805} & \underline{0.803} & 0.822 & \underline{0.804} & \underline{0.815} & \textbf{0.870} & \underline{0.857} & \textbf{0.848} & \textbf{0.880} & \textbf{0.879} & \textbf{0.826} \\
SSIM ($NeuS$) & 0.732 & \underline{0.778} & 0.722 & 0.739 & \textbf{0.915} & \textbf{0.809} & \textbf{0.818} & \textbf{0.831} & \textbf{0.812} & \underline{0.815} & \underline{0.866} & \textbf{0.863} & \underline{0.847} & \underline{0.878} & \underline{0.878} & \underline{0.820} \\
SSIM ($RF$) & \underline{0.796} & 0.746 & \textbf{0.807} & \textbf{0.780} & \underline{0.890} & 0.770 & 0.765 & \underline{0.824} & 0.727 & \textbf{0.827} & 0.845 & 0.851 & 0.808 & 0.859 & 0.846 & 0.809 \\
SSIM ($SDFoam$) & \textbf{0.805} & 0.758 & 0.744 & 0.709 & 0.858 & 0.774 & 0.748 & 0.778 & 0.755 & 0.706 & 0.813 & 0.798 & 0.790 & 0.790 & 0.750 & 0.772 \\
\hline
\end{tabular}
\caption{Quantitative comparisons with NeRF, NeuS and RF on the task of novel view synthesis without mask supervision.}
\end{table*}

\begin{table*}[h]
\centering
\scriptsize
\begin{tabular}{l|cccccccccccccccccc|c}
\hline
\textbf{Scan ID} & 24 & 37 & 40 & 55 & 63 & 65 & 69 & 83 & 97 & 105 & 106 & 110 & 114 & 118 & 122 & Mean \\
\hline
PSNR ($RF_M$) & 24.36 & 22.39 & 26.34 & \textbf{26.77} & 28.38 & 24.81 & 22.96 & 28.33 & 21.33 & 27.03 & 23.72 & 24.21 & 24.78 & 25.93 & 28.84 & 25.34 \\
PSNR ($SDFoam_M$) & \textbf{25.59} & \textbf{24.96} & \textbf{26.55} & 26.27 & \textbf{28.81} & \textbf{25.78} & \textbf{23.77} & \textbf{28.90} & \textbf{26.12} & \textbf{28.30} & \textbf{24.08} & \textbf{25.80} & \textbf{27.13} & \textbf{27.58} & \textbf{29.29} & \textbf{26.60} \\
\hline
SSIM ($RF_M$) & 0.796 & 0.770 & \textbf{0.823} & \textbf{0.925} & \textbf{0.942} & 0.901 & \textbf{0.880} & \textbf{0.952} & 0.839 & \textbf{0.917} & \textbf{0.892} & \textbf{0.907} & 0.866 & \textbf{0.927} & 0.938 & 0.885 \\
SSIM ($SDFoam_M$) & \textbf{0.803} & \textbf{0.828} & 0.814 & 0.900 & 0.937 & \textbf{0.911} & 0.870 & 0.949 & \textbf{0.904} & 0.914 & 0.886 & 0.904 & \textbf{0.884} & 0.922 & \textbf{0.939} & \textbf{0.891} \\
\hline
\end{tabular}
\caption{Quantitative comparisons with RadiantFoam on the task of novel view synthesis with mask supervision. Our joint Voronoi-SDF formulation acts as an additional regularization to boost visual appearance, removing occasional floaters in the scene.}
\end{table*}

\begin{figure*}
    \centering
    \includegraphics[width=\textwidth]{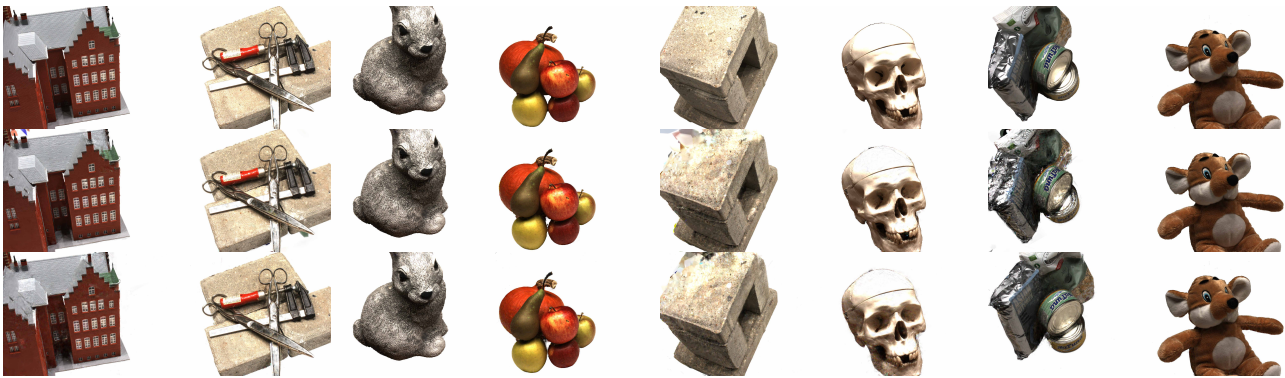}
    \captionof{figure}{Novel view synthesis qualitative results. Top to bottom: ground truth, Radiant Foam, SDFoam. Our method is able to better model reflections, as can be seen in the metallic examples, has less floaters, while retaining a very good visual fidelity in highly textured surfaces (fur, stone, scratches, etc.).}
    \label{fig:nvs}
\end{figure*}

\section{Experiments}
\label{sec:experiments}

All the experiments were run on two NVIDIA RTX PRO 6000 Blackwell Max-Q Workstation Edition GPUs granted by NVIDIA through the Academic Grant Program\footnote{\url{https://www.nvidia.com/en-us/industries/higher-education-research/academic-grant-program/}}.



We evaluate SDFoam on the DTU dataset, following the standard evaluation protocol used in prior work on neural rendering and SDF-based reconstruction. We compare against NeRF, NeuS, IDR, and RadiantFoam, using Chamfer distance, PSNR, and SSIM as quantitative metrics, as reported in \cite{wang2021neus}, leaving out 10\% of the images in the novel view synthesis task. RadiantFoam serves as our closest baseline, since it also relies on a Voronoi/Delaunay scene structure but does not model an explicit signed distance field.

\paragraph{Geometry reconstruction.}
Table~1 reports masked Chamfer distances for all the methods together with PSNR and SSIM for RF and SDFoam (in this scenario the entire dataset is used for scene reconstruction). NeuS achieves the lowest Chamfer distance on average, but both training and mesh extraction are slower on NeuS (up to 3$\times$ combined), while we simplify the SDF formulation by making it piece-constant for each cell. RadiantFoam, instead, suffers from large Chamfer errors (\emph{RF*} in Table~1) due to the presence of many floaters and the difficulty of removing them using simple density thresholding. In contrast, SDFoam reduces the average Chamfer distance of RF by more than half, while preserving competitive image quality. 

Qualitative results in Fig.~\ref{fig:mesh_qualitative} highlight the geometric advantages of our formulation. Surfaces reconstructed by RadiantFoam tend to exhibit small holes and local inconsistencies induced by ray-tracing artifacts and residual floaters. By interpreting each Voronoi cell as a local SDF, SDFoam yields meshes that are noticeably cleaner and more complete, with fewer holes and significantly fewer isolated floaters. Moreover, the resulting mesh is a \emph{surface} mesh with an empty interior, rather than a volumetrically filled mesh as in RadiantFoam, which simplifies downstream geometric processing. 

\paragraph{Novel view synthesis.}
Tables~2 and~3 summarize novel view synthesis performance without and with mask supervision, respectively. Without masks (Table~2), SDFoam attains PSNR and SSIM that are in the same range as RadiantFoam across all scans, confirming that introducing an explicit SDF does not degrade visual fidelity. When mask supervision is available (Table~3), the masked variant $SDFoam_M$ consistently outperforms $RF_M$ in both PSNR and SSIM, with an average gain of about $+1.3$ dB PSNR and a small but consistent improvement in SSIM. 

Figure~\ref{fig:nvs} illustrates qualitative novel view synthesis results. SDFoam produces renderings with visual quality comparable to RadiantFoam, while exhibiting fewer floaters and cleaner object silhouettes.

\paragraph{Runtime and mesh extraction.} Because SDFoam shares the same underlying Voronoi/Delaunay scene structure as RF and differs mainly in how each cell is parameterized and rendered, training speed is comparable to RF for a given dataset and resolution. The main practical advantage of our representation appears in the mesh extraction stage. RadiantFoam operates on a volumetrically filled mesh, which makes high-resolution isosurface extraction computationally demanding. In SDFoam, iso-surface extraction is performed directly on the SDF-conditioned Voronoi cells, producing a surface-only mesh and avoiding unnecessary processing of the interior volume. This improvement stems from using a density conversion similar to \cite{wang2021neus}, where density is defined as the derivative of the sigmoid. In our experiments, this leads to up to a $5\times$ speed-up in mesh reconstruction time compared to RadiantFoam, while at the same time delivering cleaner, more complete surfaces as evidenced by both Chamfer distance and visual inspection. 


\section{Conclusions} 
\label{sec:conclusions}

We introduced \textbf{SDFoam}, a unified representation that couples a learnable signed distance field with a Voronoi-Delaunay scene structure. By treating each Voronoi cell as a local SDF, our method brings implicit geometry into an explicit, ray-traceable formulation. This hybrid design yields geometry that is significantly cleaner than RadiantFoam while preserving comparable novel view synthesis quality.

A key strength of SDFoam is its efficient and topology-preserving mesh extraction pipeline. Instead of operating on a volumetrically filled representation, SDFoam recovers a surface-only mesh directly from SDF-conditioned Voronoi cells, while being much faster. Training speed remains on par with RadiantFoam, demonstrating that improved geometry does not come at the cost of photometric fidelity or efficiency.

Overall, SDFoam bridges the gap between radiance-driven and SDF-based methods, offering a compelling trade-off between appearance and geometry. 

While SDFoam offers a strong balance between appearance and geometry, it still has a few limitations. The Voronoi tessellation relies on a fixed number of sites, which may struggle to capture very thin or highly detailed structures and could benefit from more adaptive refinement. Our NeuS-based density mapping sharpens surfaces effectively, but its dependence on the parameter $\beta$ can make training sensitive, especially with sparse input views. Finally, although mesh extraction is much faster than RadiantFoam, it still depends on SDF thresholding, which may affect surface completeness and watertightness.

In future work, we plan to explore fully adaptive Voronoi refinement strategies, improved SDF conditioning, and integrating learning-based priors for both appearance and geometry. Extending SDFoam to dynamic scenes, non-Lambertian materials, or large-scale environments also represents an exciting direction for further investigation.

\paragraph{Acknowledgements} This research was supported by grants from NVIDIA and utilized two NVIDIA RTX PRO 6000 Blackwell Max-Q Workstation Edition GPUs.
{
    \small
    \bibliographystyle{ieeenat_fullname}
    \bibliography{main}
}

\onecolumn
\clearpage
\setcounter{page}{1}
\maketitlesupplementary

\begin{figure}
    \centering
    \includegraphics[width=\linewidth]{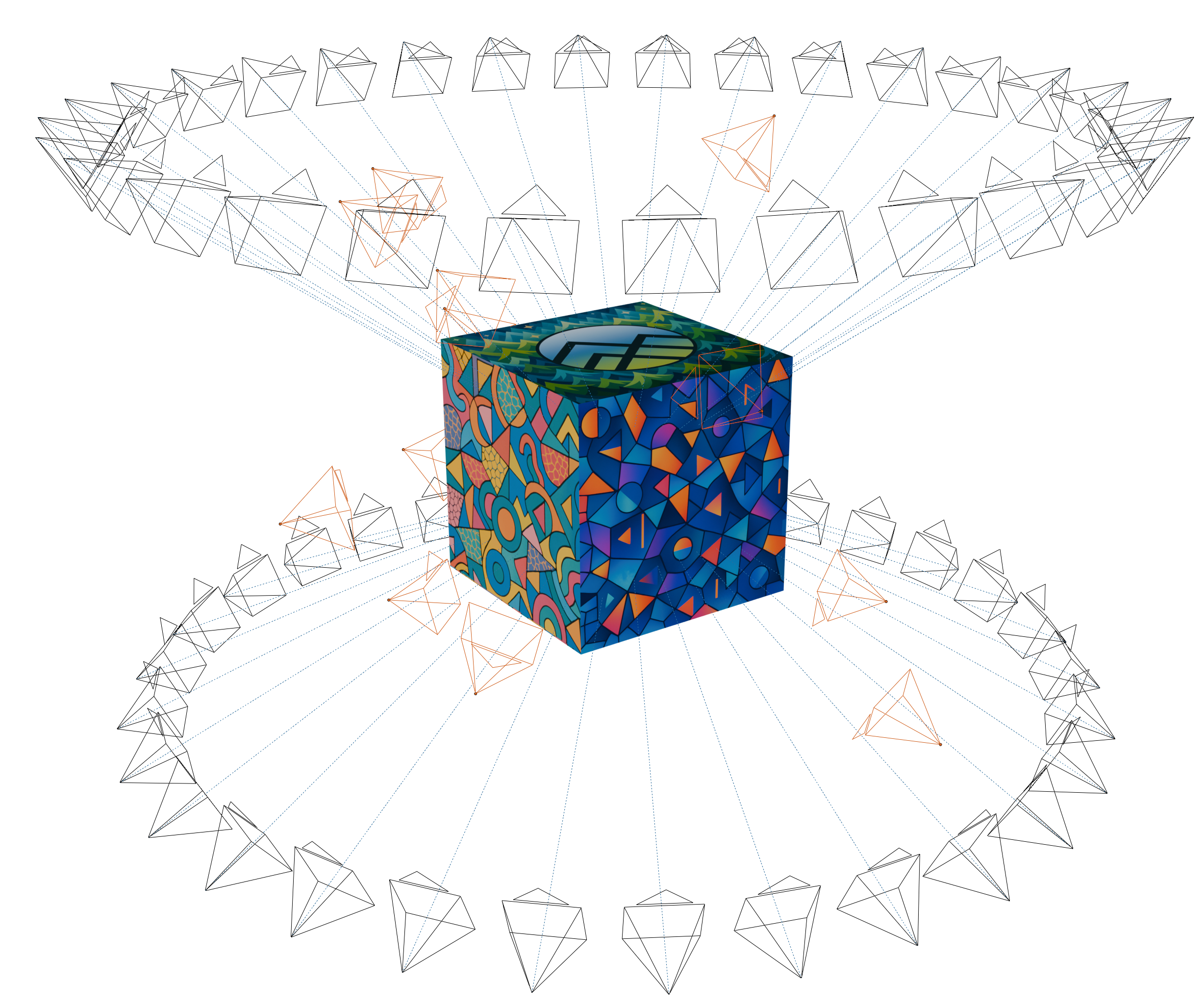}
    \caption{Synthetic scene made with Blender. The cameras placed on the two parallel circumferences (in black) are used as the training set, while the others (in orange) form the test set.}
    \label{fig:blender}
\end{figure}



\section{Synthetic 3D reconstruction benchmark}

To evaluate the ability of our method to recover clean and watertight surfaces, we construct a controlled synthetic scene in Blender consisting of a textured cube observed by 72 calibrated cameras (60 used for training and 12 for testing), as shown in Fig.~\ref{fig:blender}. We train both RadiantFoam (RF) and our SDFoam model using identical hyperparameters and camera configurations to ensure a fair, one-to-one comparison.

This setup exposes a characteristic failure scenario of RF: despite the simplicity of the underlying geometry, the reconstructed density field often develops discontinuities and \emph{holes} even on perfectly planar surfaces (Fig. \ref{fig:comparison}). This issue arises because RF represents density as an independent trainable parameter per Voronoi cell, with no enforced consistency across cell boundaries.

In contrast, SDFoam leverages a continuous signed distance field to impose geometric coherence across the entire scene. Rather than learning density directly, we compute it as the derivative of a sigmoid applied to the SDF, modulated by a learnable sharpness parameter, as defined in Eqs.~\ref{eq:sdf_density_math} and~\ref{eq:sigma}. This formulation produces a smooth, structurally consistent density field and prevents the surface fragmentation observed in RF.

\begin{figure}
    \centering
    \includegraphics[width=\linewidth]{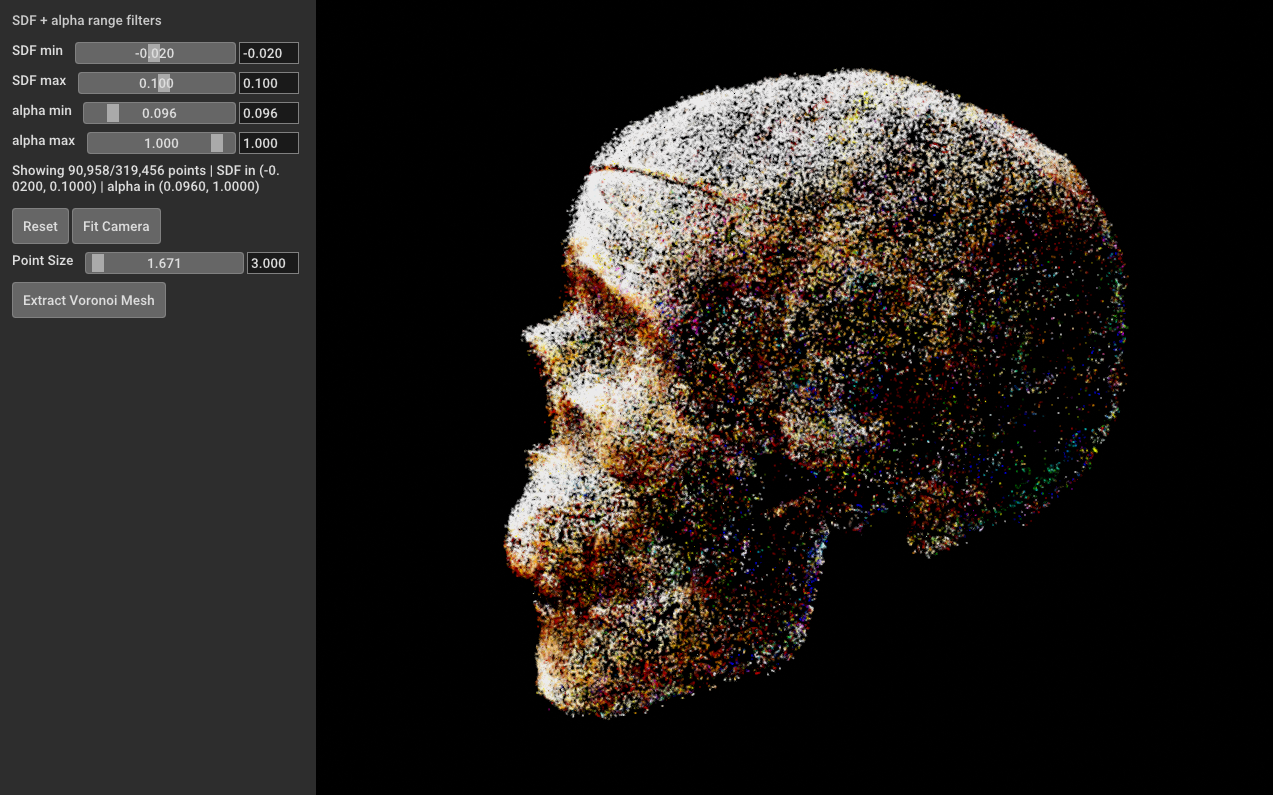}
    \caption{SDFoam GUI. The voronoi seeds can be filtered using a combination of a SDF threshold and an alpha threshold, and the geometry can be computed by obtaining the corresponding Voronoi vertices and faces from the Voronoi diagram.}
    \label{fig:GUI}
\end{figure}

\section{Surface flatness results}
Figure \ref{fig:comparison} reports additional qualitative results that highlight the behavior of the two implicit surface representations. We visualize the depth fields produced by RF (RadiantFoam) and by our SDFoam model. As shown, SDFoam is able to recover a consistent and complete geometry even in the presence of complex textures, whereas RF struggles to maintain geometric coherence and often produces holes or incomplete surfaces. This issue is inherited from the original NeRF formulation, where geometry is only indirectly induced and becomes unstable in texture-rich regions. In contrast, the explicit SDF formulation in SDFoam produces more robust and stable reconstructions while preserving a PSNR that remains quantitatively and qualitatively comparable to RF.

\begin{figure*}
    \centering
    \includegraphics[width=0.49\linewidth]{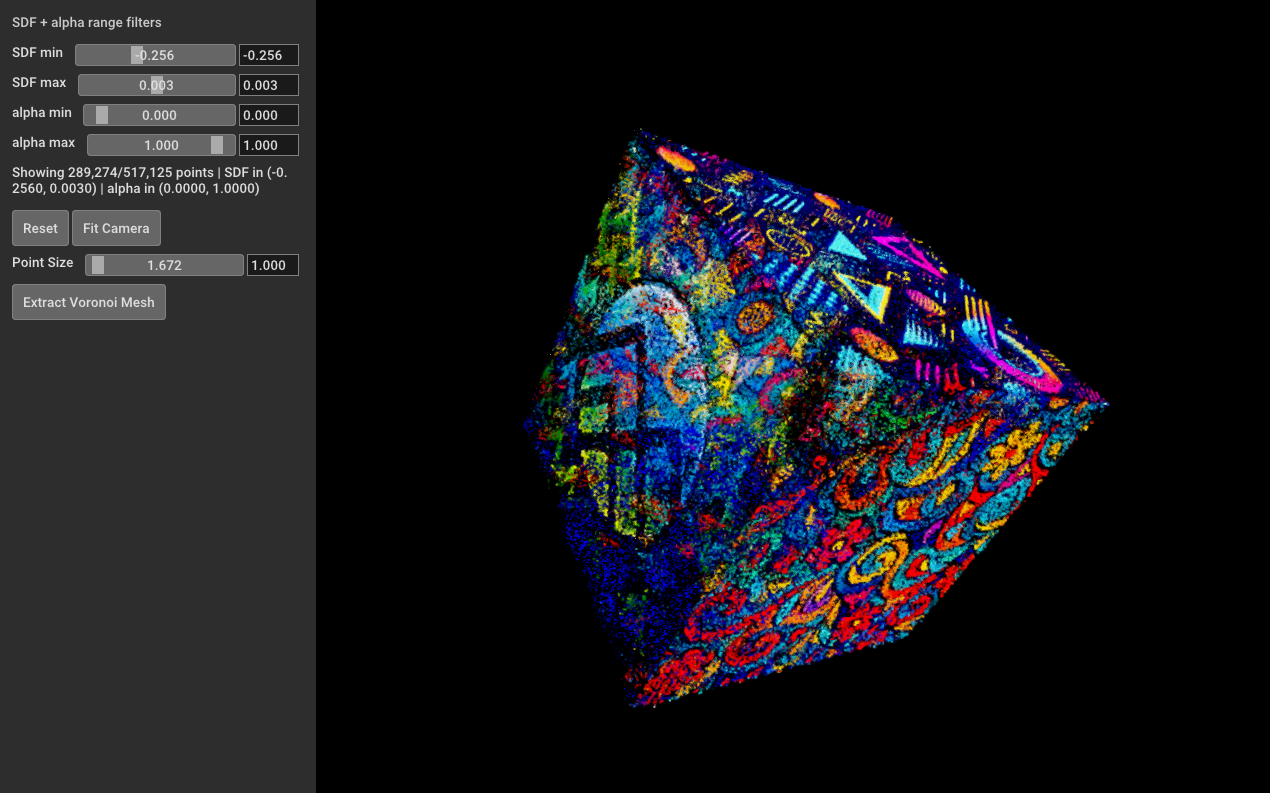}
    \includegraphics[width=0.49\linewidth]{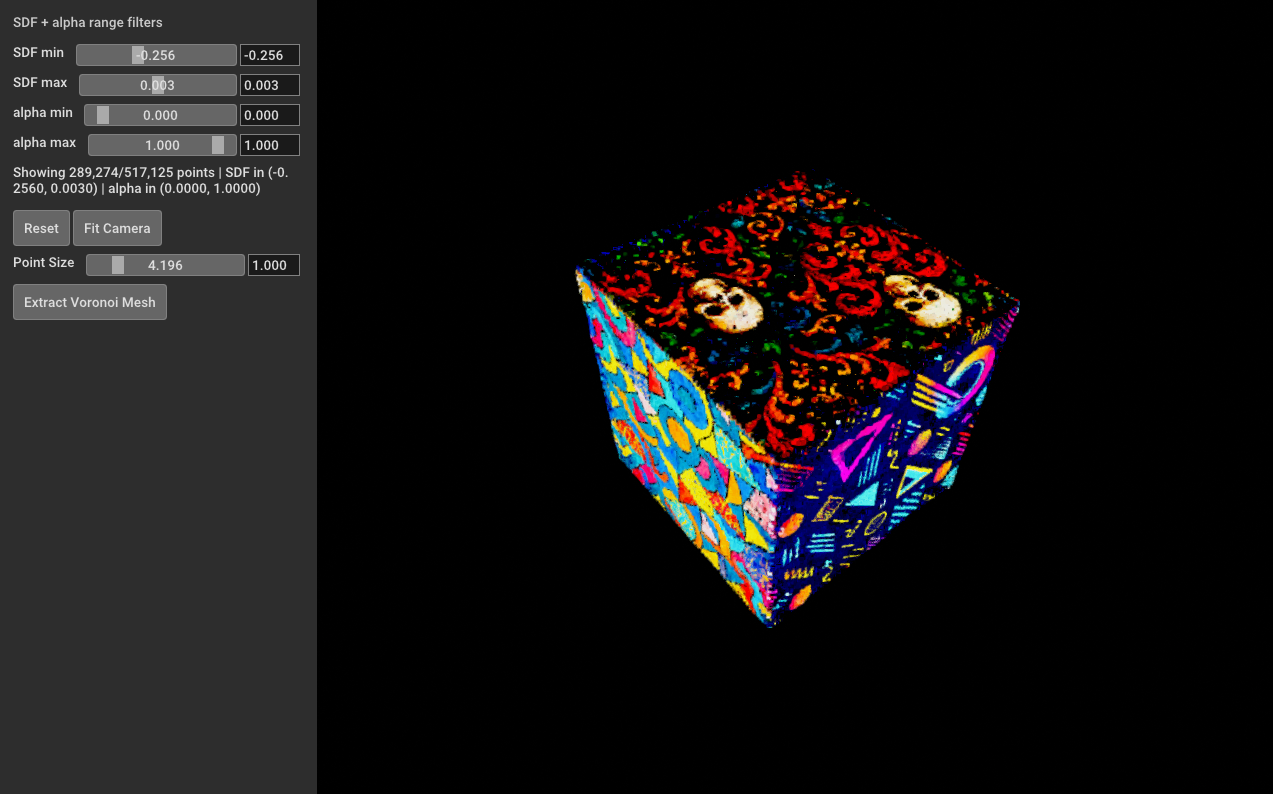}
    \caption{\textbf{SDFoam GUI}. Left: filtering the Voronoi sites through alpha + SDF thresholding. Right: the output extracted textured mesh.}
    \label{fig:seeds_to_mesh}
\end{figure*}

\section{SDFoam GUI}
We developed an interactive GUI (Figure \ref{fig:GUI}) that enables loading a trained scene and visualizing all the Voronoi sites together with their associated colors. The GUI provides fine-grained control over the selection of the sites that contribute to the final geometry. In particular, each site is associated with an SDF value, which we obtain by querying the trained MLP at the site position, and with an alpha value representing its opacity. The interface therefore exposes two independent filtering modules: one operating on the SDF range and one operating on the alpha range.

The user can select a minimum and maximum threshold for both quantities, and a Voronoi site is retained only if it satisfies both conditions simultaneously. This dual filtering mechanism is crucial, since SDF alone is often insufficient to isolate clean geometric structures. By fine-tuning the two thresholds jointly, the user can interactively refine the subset of Voronoi cells that correspond to the actual surface of the reconstructed object. Fig~\ref{fig:seeds_to_mesh} shows an example. 

\begin{figure*}
    \centering
    \includegraphics[width=\linewidth]{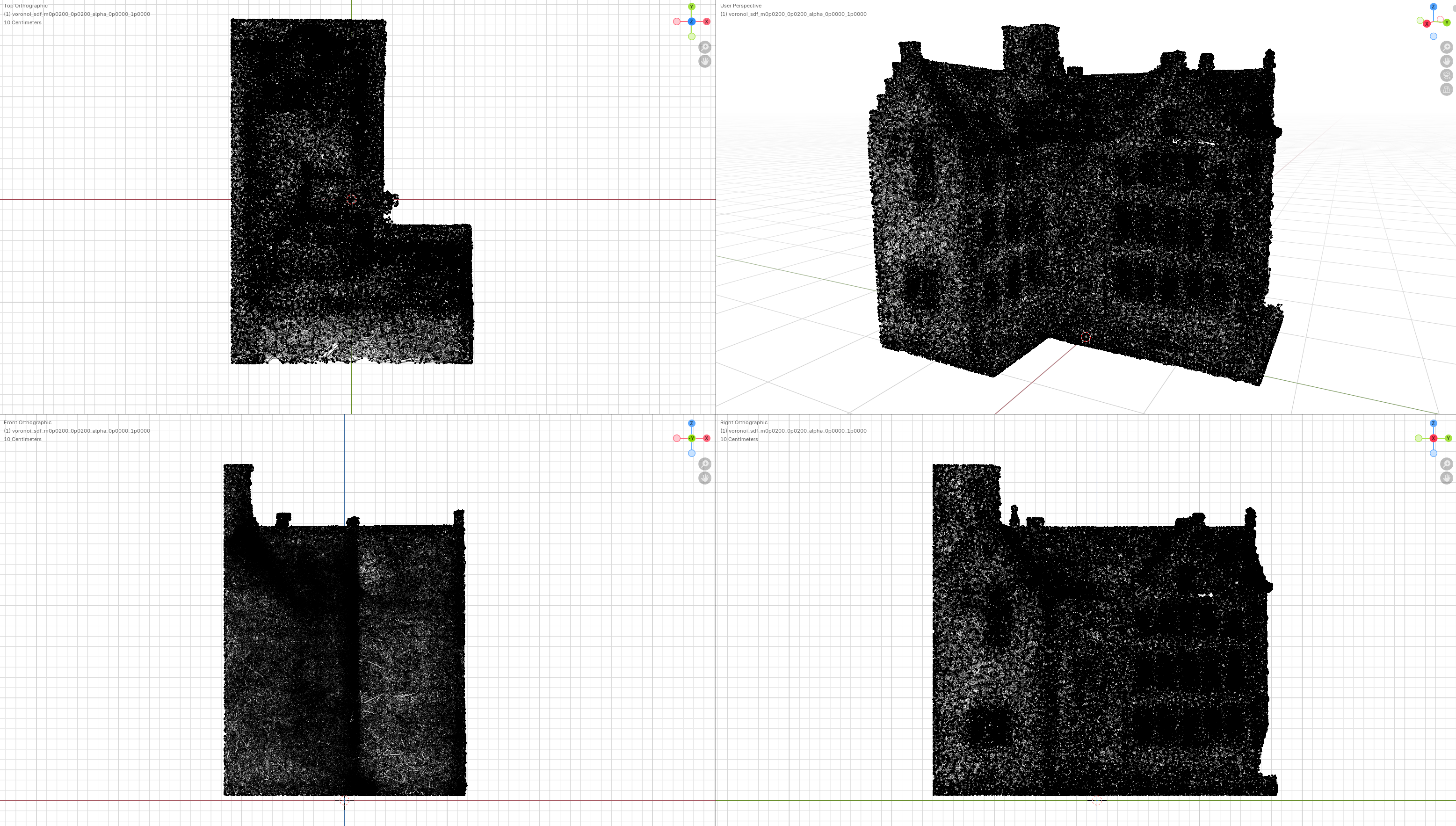}
    \caption{Loading the SDFoam extracted mesh into Blender or any other software for further processing. As seen from the wireframe orthographic views, the walls of the building remain straight, and the high poly count allows each face to retain a sigle color, which is needed for visual fidelity. A lower polygon count mesh can be obtained at this stage through remeshing and uv remapping. }
    \label{fig:blender_house}
\end{figure*}

Once a satisfactory subset of sites has been selected, the GUI provides a one-click tool to explicitly compute the full Voronoi diagram restricted to the retained cells. For each site, we compute the corresponding Voronoi region, extract its polygonal faces, and identify all its vertices, producing an explicit polygonal description of the diagram. The resulting mesh is immediately displayed in the GUI’s 3D viewport, allowing the user to visually inspect the reconstructed geometry. Additionally, the mesh can be exported as a standard mesh file, as seen in Fig. \ref{fig:blender_house}. 

\section{Mesh Extraction}
\begin{figure}
    \centering
    \includegraphics[width=0.49\linewidth]{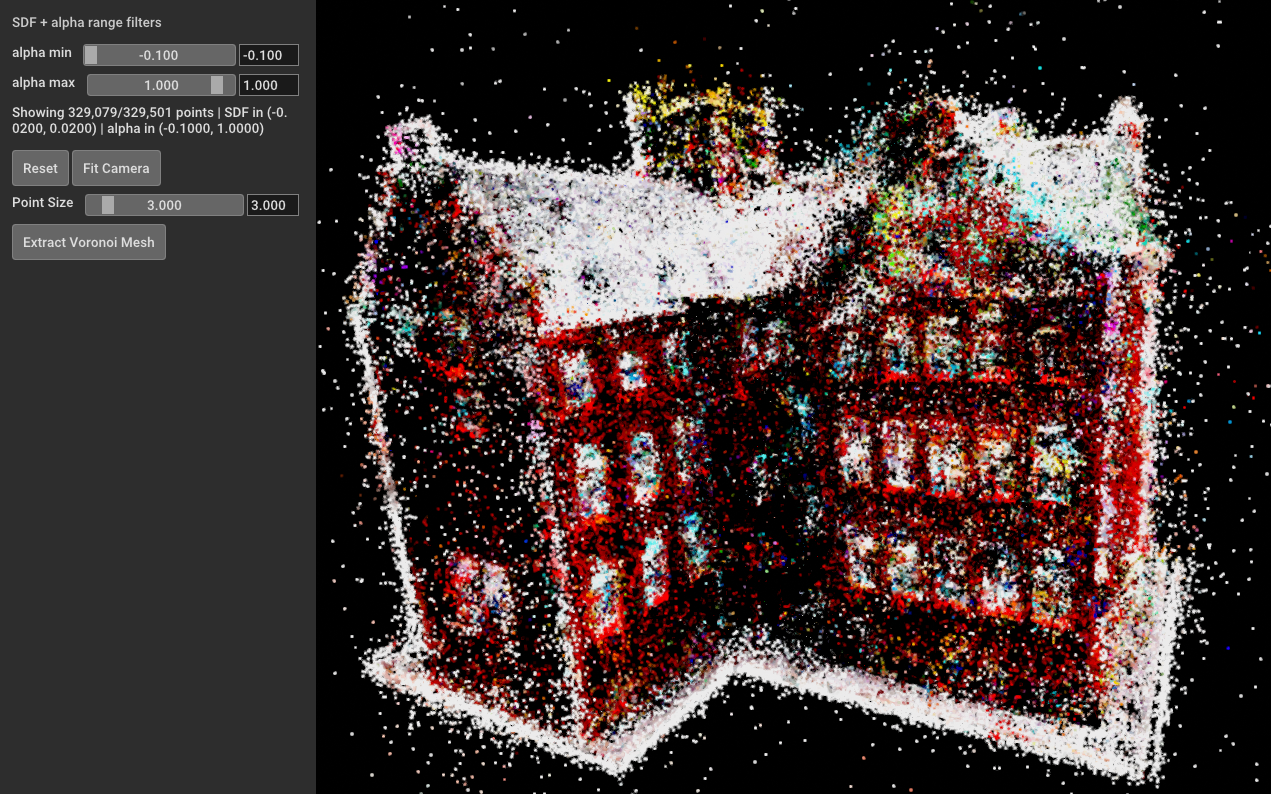}
    \includegraphics[width=0.49\linewidth]{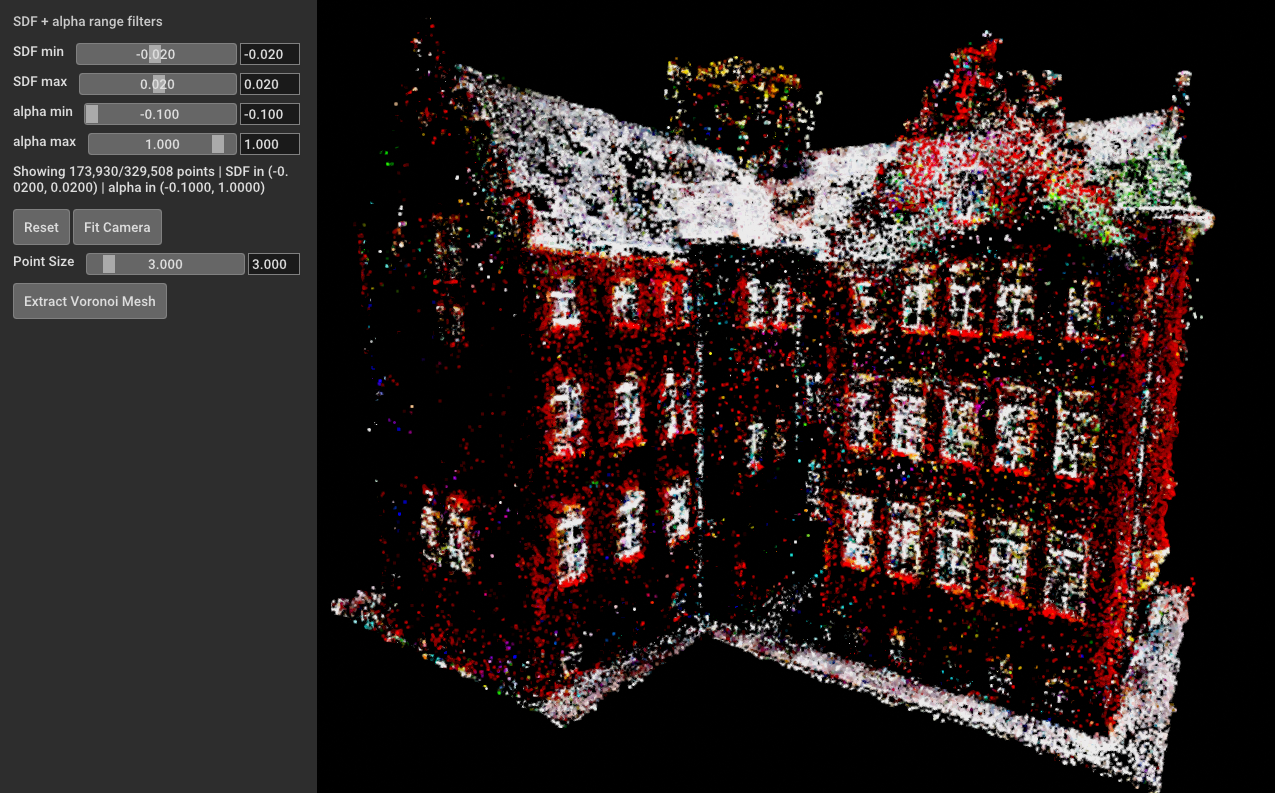}
    \caption{Left: RF, Right: SDFoam. On RF we can filter out cells based on their alpha value. However, sometimes thresholding is not enough and floating or unwanted sites remain. On SDFoam, we can filter sites both by their alpha value and SDF value, precisely removing any unwanted site. The processed SDFoam scene can be left as is, or converted into a colored point cloud or mesh.}
    \label{fig:versus}
\end{figure}

\begin{figure*}
    \centering
    \includegraphics[width=.85\linewidth]{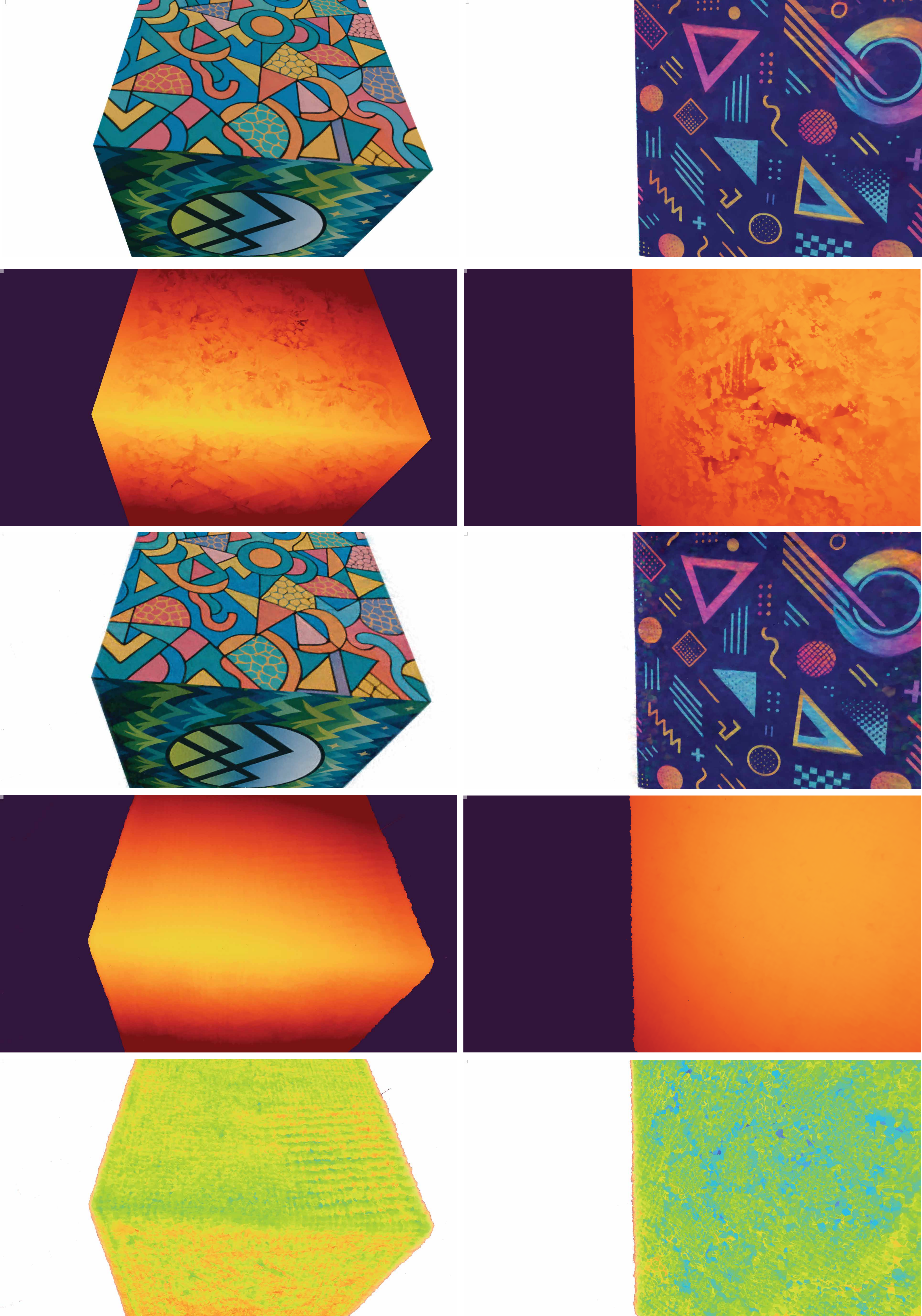}
    \caption{Qualitative comparison of geometry and viewpoint rendering. From top to bottom: RadiantFoam RGB rendering, RadiantFoam depth, SDFoam RGB rendering, SDFoam depth, and (last row) SDFoam per-cell SDF.}
    \label{fig:comparison}
\end{figure*}

We also report qualitative results of the meshes extracted from the SDF field shown in Fig.~\ref{fig:comparison}. Figure~\ref{fig:versus} compares the output of our method with the one extracted from RF. In the case of RF, the mesh is obtained through a naïve procedure. First, we assign a label to each site based on its density value, which allows us to identify the cells belonging to the object. Then, for every cell, we select the faces that are shared with adjacent cells whose density differs significantly. These face discontinuities are then used to extract the final mesh. In other words, we retain only the faces where the density contrast between neighboring cells is high (where the labels differ) and treat these as the polygonal faces of the resulting mesh. In contrast, our method extracts the surface directly from the SDF field by applying a threshold that can be tuned depending on the scene. As shown on the left of  Fig.~\ref{fig:seeds_to_mesh}, this threshold can be adjusted interactively through the control panel. The extracted geometry already retains the color information: during extraction, each Voronoi cell is assigned the color of its corresponding site. This naturally produces a texture, since the color information is inherently encoded in the Voronoi representation. An example of the extracted mesh of the cube, with the texture applied by SDFoam GUI, is shown on the right of Fig.~\ref{fig:seeds_to_mesh}. 

\begin{figure}
    \centering
    \includegraphics[width=0.4\linewidth]{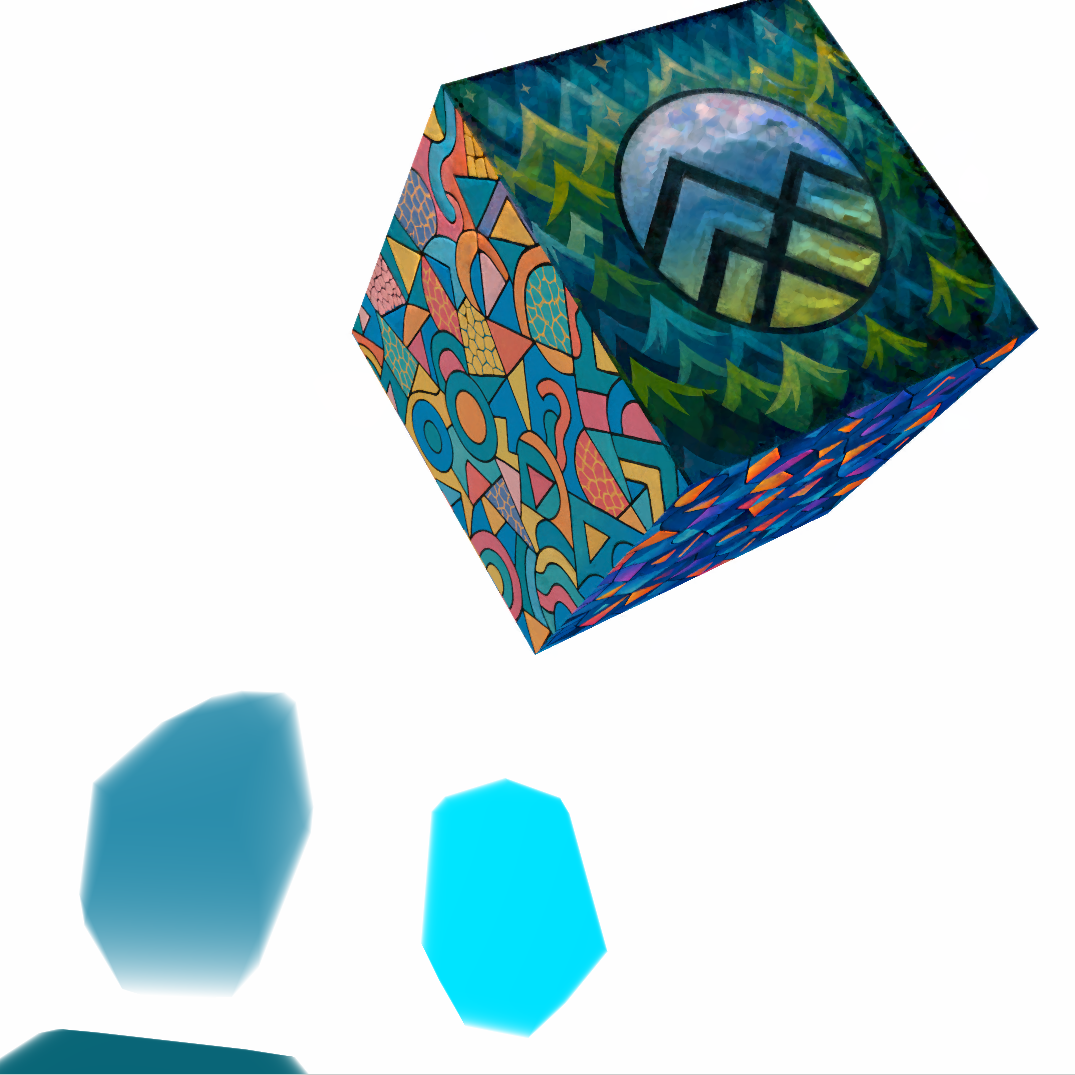}
    \includegraphics[width=0.4\linewidth]{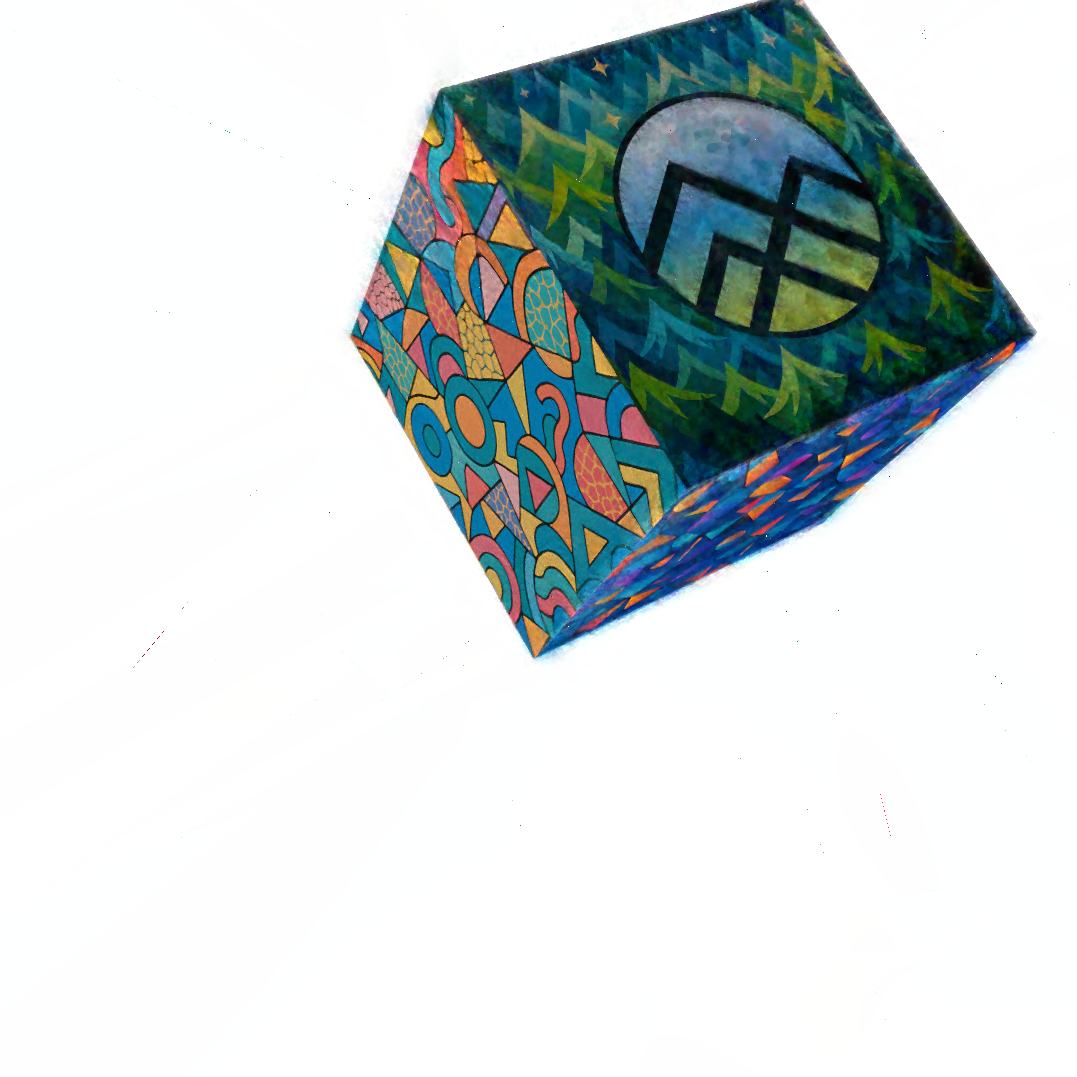}
    \caption{Left: RF, Right: SDFoam. Our method successfully gets rid of floaters by leveraging the per-cell SDF values and converting them to density.}
    \label{fig:floaters}
\end{figure}

\end{document}